\documentclass[runningheads]{llncs}

\usepackage[T1]{fontenc}
\usepackage{graphicx}
\usepackage{amsmath}
\usepackage{amsfonts}
\usepackage{amssymb}
\usepackage{booktabs}
\usepackage{multirow}
\usepackage{bbding}         
\usepackage{hyperref}    
\usepackage{color}

\makeatletter
\@ifundefined{discintname}{\def\discintname{Disclosure of Interests.}}{}
\@ifundefined{credits}{%
  \newenvironment{credits}{\begingroup\small}{\par\endgroup}}{}
\makeatother
\begin{document}
\title{JAL-Turn: Joint Acoustic--Linguistic Modeling for Real-Time and Robust
Turn-Taking Detection in Full-Duplex Spoken Dialogue Systems}

\titlerunning{JAL-Turn: Real-Time and Robust Turn-Taking Detection}

\author{%
Yu Pan\orcidID{0009-0000-4995-0689}\thanks{Equal contribution.} \and
Guangzhao Yang\orcidID{0009-0006-8520-9824}\textsuperscript{$\star$} \and
Muhammad Huzaifah\orcidID{0000-0002-7188-3600} \and
Ningjie Bai\orcidID{0009-0004-3257-8488} \and 
Shi Qiu\orcidID{0009-0001-3533-6239}\,\textsuperscript{(\Envelope)}%
\thanks{Corresponding author.} 
}

\authorrunning{Y. Pan et al.}

\institute{Recho, Inc., Tokyo, Japan\\
\email{\{y.pan,g.yang,m.huzaifah,n.haku,s.qiu\}@recho-ai.com}}

\maketitle              
\begin{abstract}
Efficient and robust turn-taking detection remains a significant challenge in industrial-grade voice AI agents. Existing systems often rely on acoustic or linguistic cues alone, limiting accuracy and stability, whereas directly endowing large language models with full-duplex capability incurs heavy data and deployment costs that hinder real-time usage. We propose JAL-Turn, a lightweight speech-only turn-taking framework based on a joint acoustic–linguistic modeling paradigm, in which a cross-attention module adaptively fuses pre-trained acoustic and linguistic representations for low-latency hold/shift prediction. By sharing a frozen ASR encoder, JAL-Turn enables turn-taking prediction to run fully in parallel with speech recognition, introducing almost no additional end-to-end latency. We further introduce a scalable pipeline that automatically derives reliable turn-taking labels from large-scale real-world dialogue corpora. Experiments on the public Easy-Turn and Smart-Turn benchmarks and an in-house Japanese customer-service dataset show that JAL-Turn achieves competing results among strong baselines in detection accuracy while maintaining superior real-time performance.

\end{abstract}
\section{Introduction}

The rapid proliferation of voice AI agents in applications such as intelligent customer service, personal assistants, and human–AI collaboration has raised the demand for natural and high-quality spoken interaction. Designed for extremely low-latency responses and continuous listening, these agents are prone to intervening before users have finished speaking. Yet natural speech is full of thinking pauses, hesitations, and self-repairs that do not indicate turn completion, and overly aggressive responses cause interruptions that disrupt conversational flow and erode user trust. Therefore, accurately determining whether a user has genuinely finished their turn—while remaining responsive—is a critical challenge for building effective, user-friendly voice AI agents.

Turn-taking is a fundamental property of human spoken interaction \cite{skantze2021turn}, in which speaking turns are exchanged naturally with minimal delay. In contrast, due to the inherent variability of the speech signal \cite{pan2024gmp,inoue2024real,inoue2025yeah}, achieving similarly real-time and stable turn management remains challenging. Although existing turn-taking detection approaches \cite{gu2024positive,skantze2025applying} have made notable progress, they still yield erroneous decisions that cause excessive response latency, frequent interruptions, and unnatural interaction dynamics, underscoring the need for real-time and robust turn-taking techniques.
Traditional spoken dialogue systems typically rely on heuristic, silence-based strategies \cite{skantze2021turn}, waiting for a fixed or adaptive silence duration before deciding that the user's turn has ended. However, silence alone is an unreliable cue: users frequently produce within-utterance pauses that do not signal a handover \cite{majlesi2023managing,inoue2024real}. Previous studies \cite{stivers2009universals,inoue2024multilingual} show that turn transitions in many languages are rapid, often on the order of 100–500~ms, indicating that listeners do not merely react to silence but proactively anticipate turn completions using lexical, prosodic, and multimodal cues.

To this end, several works build data-driven detectors from linguistic\footnote{TEN Turn Detection: https://github.com/ten-framework/ten-turn-detection; Smart-Turn (text): https://github.com/pipecat-ai/smart-turn/tree/main} or acoustic\footnote{Smart-Turn (audio): https://github.com/pipecat-ai/smart-turn/tree/filipi/smart-turn; VAP-Realtime: https://github.com/inokoj/VAP-Realtime} features; under stringent real-time constraints, however, they adopt simple architectures that limit fine-grained discrimination and leave room for improvement in accuracy and robustness. Motivated by the success of large language models (LLMs) \cite{bai2023qwen,achiam2023gpt} and speech language models (SLMs) \cite{fang2024llama,pan2025s2st}, another line \cite{defossez2024moshi,yu2024salmonn,zhang2025omniflatten,li2025easy} endows LLM/SLM backbones with full-duplex capability. Despite higher quality, such systems are hard to deploy: they require large amounts of costly manual annotation that is difficult to scale across domains and languages; their backbones must also serve other functions such as automatic speech recognition (ASR), adding latency \cite{li2025easy}; and their semantic emphasis discards fine-grained acoustic cues crucial for turn-taking, degrading accuracy in complex real-world scenarios.

In this paper, we propose \textbf{JAL-Turn}, a lightweight speech-only turn-taking model that jointly leverages pre-trained acoustic and linguistic encoders and supports parallel inference with ASR through encoder sharing, enabling low-latency deployment. Additionally, we introduce a scalable data construction pipeline that automatically derives reliable turn-taking labels from large-scale real-world dialogue corpora without manual annotation, enabling effective training across domains and languages. 
Experiments on the public Easy-Turn and Smart-Turn benchmarks and an in-house Japanese customer-service corpus show that JAL-Turn outperforms strong audio-only and LLM-based baselines in detection accuracy at latency suitable for real-time deployment. Ablation studies and attribution analyses further confirm the effectiveness of the proposed approach.

\begin{figure}[t]
\centering
    \includegraphics[width=\linewidth]{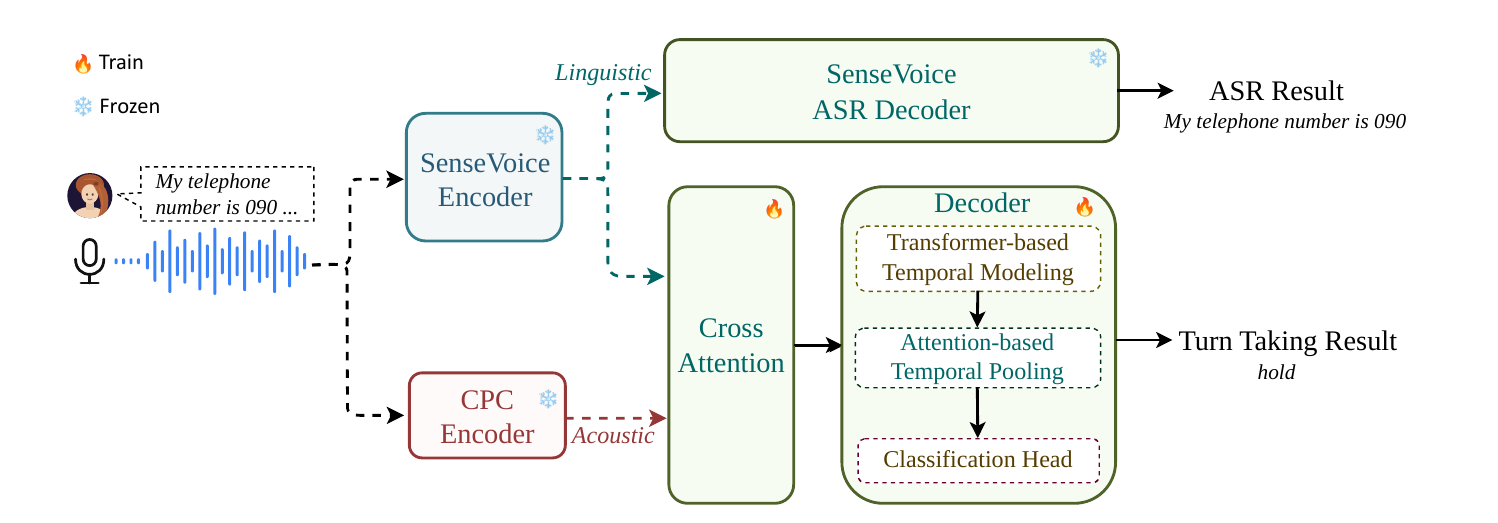}
    \caption{Overall training architecture of the proposed JAL-Turn framework.}
    \label{fig:JAL-Turn}
\end{figure}

\section{Related Work}

\noindent\textbf{Turn-taking detection.} Early spoken dialogue systems rely on heuristic silence-based endpointing~\cite{skantze2021turn}, which is brittle to within-utterance pauses. Voice Activity Projection (VAP)~\cite{ekstedt2022voice,inoue2024real,inoue2025yeah} instead predicts upcoming voice-activity patterns from acoustic input as a continuous proxy for turn dynamics. More recent data-driven detectors adopt either linguistic or acoustic cues, while a parallel line of work endows LLM/SLM backbones with full-duplex capability~\cite{defossez2024moshi,zhang2025omniflatten,li2025easy}, achieving strong quality at the cost of heavy annotation and inference overhead.

\noindent\textbf{Positioning of JAL-Turn.} The novelty of JAL-Turn lies in its real-time deployment design rather than any single module, and it differs from the two closest lines along three axes. \emph{Relative to EasyTurn model}~\cite{li2025easy}, which also fuses acoustic and linguistic cues: EasyTurn model is a generative SLM that must jointly perform ASR and turn prediction, whereas JAL-Turn is a lightweight discriminative model over two frozen encoders that shares its ASR encoder---so turn prediction runs in parallel with ASR in a single forward pass, avoiding the extra ASR-decoding and language-model inference EasyTurn incurs---and is trained purely from automatically derived labels, removing EasyTurn's manual-annotation dependence. 
\emph{Relative to VAP}~\cite{ekstedt2022voice}, which motivates our labeling, the distinction is twofold. \emph{In formulation}, VAP regresses future voice activity as its online prediction target, whereas JAL-Turn is a directly discriminative hold/shift classifier that uses the future-window rule only offline for labeling and stays strictly causal at inference. \emph{Architecturally}, VAP is audio-only, whereas JAL-Turn jointly models acoustic (CPC) and linguistic cues via cross-attention and, by sharing that encoder, produces turn decisions in parallel with ASR at almost no extra cost.

\section{JAL-Turn}

As illustrated in Fig.~\ref{fig:JAL-Turn}, the proposed \textbf{JAL-Turn} framework comprises five components: (1) a dual-path encoder for acoustic and linguistic feature extraction, (2) a cross-attention fusion module for integrating heterogeneous representations, (3) a self-attention Transformer for temporal modeling, (4) an attention pooling module for utterance-level aggregation, and (5) a binary classification head for hold/shift prediction.

\subsection{Dual-Encoder Architecture}

Unlike conventional single-encoder approaches, we adopt a dual-path encoder to capture complementary aspects of the speech signal. The primary encoder is the pretrained SenseVoice-Small \cite{an2024funaudiollm}, a multilingual speech foundation model with a convolution-augmented transformer backbone \cite{gulati2020conformer,yang2023hybridformer} that is effective at extracting phonetic, sub-lexical, and lexical linguistic information. The secondary encoder is a pretrained contrastive predictive coding (CPC) model \cite{riviere2020unsupervised} emphasizing low-level acoustic regularities via self-supervised contrastive learning.

Given an input waveform $\mathbf{x} \in \mathbb{R}^{1 \times L}$, where $L$ denotes the waveform length, the two encoders produce frame-level feature sequences:
\begin{align}
    \mathbf{h}_l &= \text{Encoder}_{\text{Sense}}(\mathbf{x}) \in \mathbb{R}^{T_1 \times d_1}, \\
    \mathbf{h}_a &= \text{Encoder}_{\text{CPC}}(\mathbf{x}) \in \mathbb{R}^{T_2 \times d_2},
\end{align}
where $d_1 = 512$ and $d_2 = 256$ are the output dimensions. Both encoders are frozen during training to preserve their pre-trained knowledge, and linear projection layers map their features into a shared latent space, letting JAL-Turn jointly exploit SenseVoice's linguistic cues (phonetic to lexical) and CPC's fine-grained acoustic patterns. Importantly, the SenseVoice encoder is shared between ASR and JAL-Turn: the turn-taking branch consumes the same shared-encoder output and runs concurrently with ASR decoding, inserting no sequential stage before or after it, so turn decisions and transcriptions are produced together.

\subsection{Cross-Attention-based Fusion}

To integrate the heterogeneous dual-path features, we introduce a cross-attention fusion module that lets JAL-Turn dynamically attend to the most informative regions across feature spaces.
Concretely, the fusion module consists of two stacked cross-attention layers. In each layer, the SenseVoice features $\mathbf{h}_l'$ act as queries, while CPC features $\mathbf{h}_a'$ serve as the keys and values.
Each layer additionally incorporates residual connections and a position-wise feed-forward network.

\subsection{Transformer-based Module}

On the fused representation, we apply a causal self-attention Transformer with Attention with Linear Biases (ALiBi) \cite{press2022train}, which replaces absolute positional embeddings with a distance-dependent bias added directly to the attention logits. ALiBi improves length extrapolation and encodes a recency bias well suited to the local temporal patterns that govern turn-taking.

\subsection{Attention-based Temporal Pooling Module}

To obtain an utterance-level representation while letting JAL-Turn focus on the most informative segments, we apply lightweight attention-based temporal pooling over the hidden states $\{\mathbf{h}_t\}_{t=1}^{T}$ produced by the Transformer module:
\begin{equation}
    \alpha_t = \operatorname{softmax}\big(\mathbf{w}^{\top} \mathbf{h}_t + b\big)
\end{equation}
\begin{equation}
   \mathbf{h}_{\text{pool}} = \sum_{t=1}^{T} \alpha_t \cdot \mathbf{h}_t
\end{equation}
where $\mathbf{w} \in \mathbb{R}^{d}$ and $b \in \mathbb{R}$ are trainable parameters, and $\alpha_t$ denotes the normalized attention weight for time step $t$.

\subsection{Classification Head and Training Objective}

Given the pooled representation $\mathbf{h}_{\text{pool}} \in \mathbb{R}^{d}$, we apply a lightweight linear classification head to produce the shift logit $\hat{y} \in \mathbb{R}$, parameterized by trainable weights $\mathbf{w}_{\text{cls}} \in \mathbb{R}^{d}$ and bias $b_{\text{cls}} \in \mathbb{R}$. The predicted probability is obtained via a sigmoid activation, and the final decision (hold vs.\ shift) is made using a fixed threshold $\theta=0.5$ (we use $\theta$ for the decision threshold to avoid collision with the future-window length $\tau$ in Eq.~(5)).
In addition, the model is trained end-to-end (E2E) with the standard binary cross-entropy objective over mini-batches, where $y \in \{0,1\}$ denotes the ground-truth turn label ($1$ for shift and $0$ for hold).

\begin{figure}[t]
\centering
    \includegraphics[width=\linewidth]{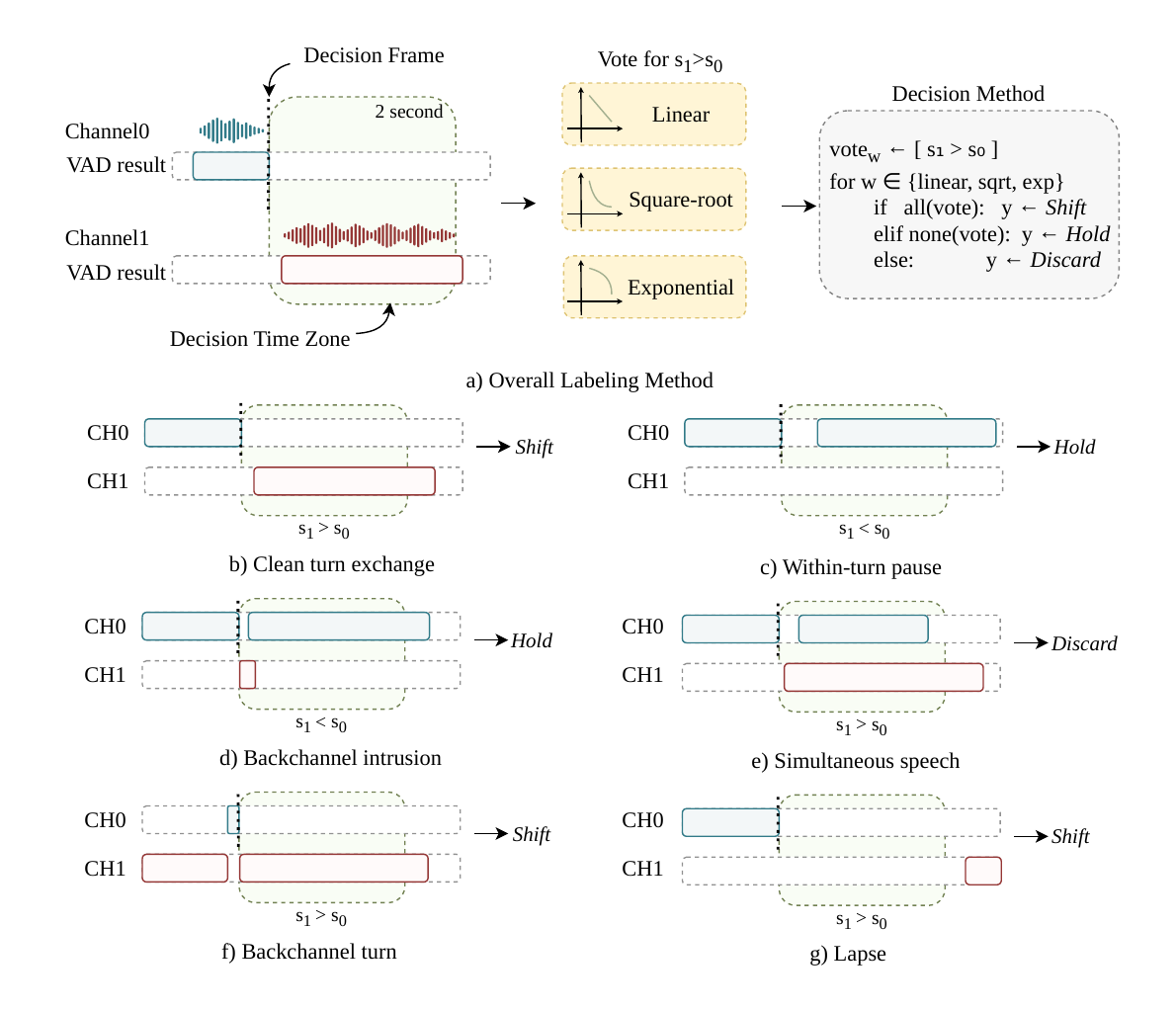}
    \caption{Schematics of the overall labeling method.}
    \label{fig:data}
\end{figure}

\section{Data Pipeline}
To enhance the robustness of our turn-taking model, we develop an efficient data pipeline that automatically derives reliable training labels from large-scale real-world conversational corpora without requiring any manual annotation. The pipeline draws on \emph{two complementary sources of supervision} that occupy opposite ends of a single trade-off: conversational recordings supply the full variability of natural turn exchanges but admit only approximate labels, whereas single-utterance recordings admit exact labels but contain no conversational dynamics. 
Crucially, both sources share the same frame-level front-end and the same context-construction back-end; they differ only in the operator that assigns Hold/Shift labels.
As shown in Fig. \ref{fig:data}, given an input recording, we first extract frame-level voice activity detection (VAD) at 50 Hz. For each channel $c \in \{0,1\}$, we obtain a binary sequence
$\mathbf{v}_c = [v_c^1, \ldots, v_c^T] \in \{0,1\}^T$,
where $v_c^t = 1$ indicates speech activity at frame $t$. This representation is common to both supervision sources described below.

\subsection{Conversational Supervision via Future-Window Labeling}
The first source consists of 1,128 hours of in-house stereo conversational data, in which both channels are active and turn exchanges arise naturally. Because the ground-truth turn boundaries are unknown, labels must be inferred from the signal itself.
Motivated by previous studies on future voice activity projection \cite{ekstedt2022voice}, we employ a future-window strategy that leverages upcoming conversational dynamics. For each frame $t$, we compute a weighted VAD score over a 2-second future window:
\begin{equation}
    s_c^t = \sum_{i=0}^{\tau f_s} w(i)\, v_c^{t+i},
\end{equation}
where $\tau = 2$ s, $f_s = 50$ Hz, and $w(i)$ is a non-negative weighting function. Three temporal weighting schemes—linear, square-root, and exponential—independently assign Hold/Shift labels (Hold if $s_c^t > s_{1-c}^t$), and a label is kept only when all schemes agree. This future-window design effectively suppresses backchannels: instantaneous VAD comparisons are easily confounded by brief listener responses, whereas future VAD patterns provide more reliable cues for genuine turn transitions.
This source yields approximately 2,299 hours of trainable segments, with an estimated labeling accuracy of about 85\% based on manual inspection. It therefore contributes conversational realism at the cost of label noise.

\subsection{Utterance-Level Supervision via Deterministic Labeling}
The second source consists of a 95-hour in-house dataset in which each recording contains a single complete utterance spoken by one speaker (e.g. ``My phone number is 090-8987-2023''). Here the turn boundary is known by construction, so no inference is required: all frames except the final one are labeled as Hold.

This source yields 749 hours of training segments. Although it achieves nearly 100\% labeling accuracy, it lacks the conversational variability present in real dialogues. It therefore contributes clean supervision at the cost of coverage—the exact converse of the conversational source.

\subsection{Context Construction}
Both sources are segmented by the same procedure. Training samples are extracted from each VAD falling edge following a speech segment, which corresponds to natural turn-taking decision points. For each such point, we construct a context window by extending backward to the previous long silence (duration $\ge 2$ seconds) and then normalize it to a fixed 10-second length through left-padding or truncation as needed, ensuring that the window always includes a complete preceding speech segment to provide sufficient semantic context for turn prediction.

\subsection{Dataset Composition}
The two sources are complementary along the accuracy--variability axis: conversational data provides large-scale turn dynamics under noisy supervision ($\sim$2,299 h, $\sim$85\% label accuracy), while utterance-level data provides high-quality supervision under restricted dynamics (749 h, $\sim$100\% label accuracy). We therefore train our model on a mixture of both datasets, combining large-scale conversational dynamics with high-quality utterance-level supervision.

\section{Experiments}
\subsection{Experimental Setups}
\label{subsec:setup}
\subsubsection{Datasets.}

To comprehensively assess the proposed method, we evaluate on the public Mandarin dataset Easy-Turn ($\approx$1145\,h), the multilingual Smart-Turn$^{3,4}$ benchmark ($\approx$700\,h across 23 languages), and the large-scale in-house corpus of real-world Japanese dialogues introduced above.
To avoid ambiguity, \emph{Easy-Turn} and \emph{Smart-Turn} always denote the public benchmark \emph{datasets}, whereas \emph{ETurn}~\cite{li2025easy} denotes the SLM-based \emph{model} released with the Easy-Turn dataset and STurn-v2/v3 denote the audio-only Smart-Turn detector models; dataset and model names are never used interchangeably.

{
\let\thefootnote\relax
\footnote{$^3$https://huggingface.co/datasets/pipecat-ai/smart-turn-data-v3-train}
\footnote{$^4$https://huggingface.co/datasets/pipecat-ai/smart-turn-data-v3-test}
}

For Easy-Turn and Smart-Turn, we evaluate on the original test sets and split the training set 9:1 into training and validation. For the in-house corpus we adopt the same 9:1 split and additionally collect 513 human-labeled samples from real-world business data as the test set, covering diverse attributes (e.g., gender, age, and business scenario) for a comprehensive evaluation.

\subsubsection{Implementation Details.}
JAL-Turn is trained E2E in PyTorch on a single H100 GPU for 10 epochs, using the AdamW optimizer (initial learning rate $1 \times 10^{-4}$, weight decay 0.001, batch size 64) with cosine-annealing decay to a minimum of $1 \times 10^{-6}$.
For a fair comparison on the Easy-Turn corpus, whose annotations distinguish four turn states (\textit{complete}, \textit{incomplete}, \textit{backchannel}, \textit{wait}), we replace JAL-Turn's binary head with a four-way head while keeping all other modules identical; on the remaining (binary) datasets we use the standard hold/shift head. 

Regarding evaluation metrics, we use accuracy and F1 to measure detection quality and latency to measure responsiveness in full-duplex interaction. Here, latency denotes the wall-clock time required to complete one forward pass of the turn-taking branch on a single H100 GPU.
Since this time scales with the effective input sequence length, it varies across datasets in proportion to their average utterance duration (approximately 3.34\,s on Easy-Turn, 7.30\,s on STurn-v3, and 7.47\,s on the in-house Japanese test set), which accounts for the 12\,ms versus 36--38\,ms difference across benchmarks. Crucially, because the turn-taking branch shares the frozen SenseVoice/ASR encoder and its decoding runs concurrently with---and is comparable in cost to---ASR decoding, this latency is effectively hidden under the ASR pipeline and introduces no additional end-to-end latency.

\subsection{Main Results}

We evaluate JAL-Turn against three categories of baselines: (1) audio-only methods (STurn-v2, STurn-v3); (2) LLM-based pipelines (GPT-5.1$^5$, Qwen3-0.6B$^6$, Gemini-2.5-Flash$^7$), where SenseVoice first produces ASR transcripts that are then fed to the LLMs for turn-state prediction; and (3) an SLM-based system, ETurn~\cite{li2025easy}, which detects turns with a speech language model backbone. GPT-5.1 and Gemini-2.5-Flash are evaluated via their official APIs, while Qwen3-0.6B is fine-tuned on the training data and served with vLLM$^{8}$~\cite{kwon2023efficient}. All LLM experiments use the same prompt (see the Appendix). Note that, unlike the on-device methods, the latencies of the API-served GPT-5.1 and Gemini-2.5-Flash include network round-trip time and are thus not measured on the same footing as the locally served GPU models; we report them to characterize practical responsiveness rather than as a strict hardware comparison, while the accuracy/F1 comparisons remain unaffected.

{
\let\thefootnote\relax
\footnote{$^5$https://openai.com/zh-Hans-CN/index/gpt-5-1}
\footnote{$^6$https://huggingface.co/Qwen/Qwen3-0.6B}
\footnote{$^7$https://poe.com/Gemini-2.5-Flash}
\footnote{$^{8}$https://github.com/vllm-project/vllm}
}

\begin{table}[ht]
\centering
\caption{Comparison of turn-taking detection methods on Mandarin Easy-Turn test set. ``-'' indicates that the state is not applicable to the given baseline.}
\label{tab:easyturn}
\renewcommand{\arraystretch}{0.8}
\begin{tabular}{lccccc}
\toprule
Model & Acc$_{\mathrm{cp}}$ & Acc$_{\mathrm{incp}}$ & Acc$_{\mathrm{bc}}$ & Acc$_{\mathrm{wait}}$ & Latency (ms) \\
\midrule
Paraformer~\cite{gao2022paraformer}\,+\,TEN Turn Detection & 86.67 & 89.30 & -              & 91.00          & 79 \\
STurn-v2                                                   & 78.67 & 62.00 & -              & -              & 16 \\
ETurn~\cite{li2025easy}                                    & 96.33 & \textbf{97.67} & \textbf{91.00} & \textbf{98.00} & 143 \\
JAL-Turn                                                   & \textbf{96.67} & 93.67 & 80.00 & 92.00 & \textbf{12} \\
\bottomrule
\end{tabular}
\end{table}


\begin{table}[ht]
\centering
\caption{Performance comparison of audio-only turn-taking detection methods on the multilingual Smart-Turn benchmark and the in-house Japanese corpus.}
\label{tab:audio-only}
\renewcommand{\arraystretch}{0.8} 
\begin{tabular}{ccccccc}
\toprule
\multirow{2}{*}{Model} & \multicolumn{3}{c}{Smart-Turn (multilingual)} & \multicolumn{3}{c}{In-house Japanese} \\
\cmidrule(lr){2-4} \cmidrule(lr){5-7}
 & Acc & F1 & Latency (ms) & Acc & F1 & Latency (ms) \\
\midrule
STurn-v2   & 65.12          & 0.679          & 48          & 55.46          & 0.427          & 51          \\
STurn-v3   & 93.10          & 0.931          & \textbf{12} & 71.94          & 0.736          & \textbf{13} \\
JAL-Turn   & \textbf{93.27} & \textbf{0.934} & 36          & \textbf{93.24} & \textbf{0.939} & 38          \\
\bottomrule
\end{tabular}
\end{table}

\subsubsection{Comparison with SLM-based Systems.}
We first compare JAL-Turn with strong baselines on the Mandarin Easy-Turn corpus (Table~\ref{tab:easyturn}), where Acc$_{cp}$, Acc$_{incp}$, Acc$_{bc}$, and Acc$_{wait}$ denote accuracy on the \textit{complete}, \textit{incomplete}, \textit{backchannel}, and \textit{wait} states, respectively (higher is better).
JAL-Turn achieves the best \texttt{cp} accuracy (96.67\%) at only 12\,ms latency, markedly outperforming Paraformer + TEN Turn Detection and STurn-v2 on the complete and incomplete states while cutting latency from 79/16\,ms to 12\,ms.
Against ETurn, JAL-Turn attains comparable performance on \texttt{incp} and \texttt{wait} (within 4.0 and 6.0 points, respectively) and slightly improves \texttt{cp} accuracy (96.67\% vs.\ 96.33\%).
However, JAL-Turn underperforms on the \texttt{bc} state (80\% vs.\ 91\%) for the following reason. As backchannels are short, semantically light responses best resolved from explicit lexical/semantic cues, yet JAL-Turn---deliberately forgoing a generative LLM/full-ASR decoding stage for low latency---accesses semantics only indirectly via the frozen ASR encoder, giving it a weaker semantic representation than ETurn's SLM backbone. Despite this gap, JAL-Turn retains a favorable overall quality--latency trade-off under strict real-time constraints.

\subsubsection{Comparison with Audio-only Methods.}

For a fair architectural comparison, STurn-v2 and STurn-v3 are re-trained from their official implementations on the in-house Japanese data under JAL-Turn's training recipe, so all audio-only systems are evaluated strictly in-domain and the gains reflect architecture rather than data mismatch. As shown in Table~\ref{tab:audio-only}, JAL-Turn attains the best accuracy and F1 on both benchmarks: on Smart-Turn it is on par with STurn-v3 (93.27\% vs.\ 93.10\%) while far exceeding STurn-v2, and on the in-house Japanese corpus the gain is far larger (93.24\%/0.939 vs.\ 71.94\%/0.736 for STurn-v3, i.e., $\sim$30\% relative accuracy and $\sim$28\% relative F1).
\begin{table}[ht]
\centering
\caption{Performance comparison of JAL-Turn against LLM-based turn-taking detection methods on the in-house benchmark.}
\label{tab:public}
\renewcommand{\arraystretch}{0.8}
\begin{tabular}{cccc}
\hline
Model & Acc & F1 & Latency (ms) \\ \hline
Gemini-2.5-Flash & 76.91 & 0.817 & 595 \\
Qwen3-0.6B & 78.70 & 0.782 & 124 \\
GPT-5.1 & 85.52 & 0.874 & 1205 \\
JAL-Turn & \textbf{93.24} & \textbf{0.939} & \textbf{38} \\
\hline
\end{tabular}
\end{table}
In efficiency, STurn-v3 has the lowest latency (12/13\,ms), but JAL-Turn stays competitive (36/38\,ms), cutting latency by about 25\% relative to STurn-v2 while delivering dramatically higher accuracy and F1. This shows that the proposed joint acoustic--linguistic paradigm yields clearly superior detection quality at low E2E latency suitable for real-time deployment.

\subsubsection{Comparison with LLM-based Methods.}
As shown in Table~\ref{tab:public}, JAL-Turn also compares favorably with LLM-based detectors on the in-house benchmark.
Compared with Gemini-2.5-Flash and Qwen3-0.6B, JAL-Turn improves accuracy by 16.3 and 14.5 percentage points and F1 by 0.122 and 0.157; against GPT-5.1, it raises accuracy from 85.52\% to 93.24\% and F1 from 0.874 to 0.939. In terms of efficiency, JAL-Turn completes each decision in 38\,ms of local inference, whereas LLM-based pipelines require full ASR decoding followed by large-scale language-model inference---incurring hundreds of milliseconds even before network overhead is considered. JAL-Turn thus matches or surpasses substantially heavier LLM pipelines in accuracy while remaining lightweight enough for real-time full-duplex interaction.

\subsection{Ablation Studies}

To assess the contribution of each component in JAL-Turn, we conduct ablation experiments on the in-house Japanese corpus, as summarized in Table~\ref{tab:ablation}.
Removing either encoder clearly degrades performance: dropping SenseVoice (w/o Sense) is most damaging (72.01\%, 0.698), confirming linguistically enriched representations as the primary driver, whereas removing CPC (w/o CPC) causes a smaller but non-trivial drop (84.18\%, 0.839) from its complementary acoustic cues. Replacing cross-attention with concatenation (w/o CrossATT, 88.59\%) shows that explicitly modeling acoustic–linguistic interactions beats simply co-presenting the two streams, and removing attention pooling (w/o ATTPooling) both lowers accuracy (90.23\%) and raises latency (38\,ms $\rightarrow$ 45\,ms), confirming that the lightweight pooling gives a better accuracy–latency trade-off than simpler aggregation.

\begin{table}[htbp]
    \centering
    \caption{Ablation studies of the proposed JAL-Turn. w/o CrossATT denotes using concatenation, w/o ATTPooling represents using the last layer feature.}
    \label{tab:ablation}
    \renewcommand{\arraystretch}{0.8}
    \begin{tabular}{cccc}
        \toprule
        Model   & Acc & F1   & Latency (ms)  \\
        \midrule
        JAL-Turn          & \textbf{93.24}  & \textbf{0.939}   & 38  \\
        w/o Sense         & 72.01           & 0.698            & 12 \\
        w/o CPC           & 84.18           & 0.839            & 26 \\
        w/o CrossATT      & 88.59           & 0.873            & 40 \\
        w/o ATTPooling    & 90.23           & 0.895            & 45 \\
        \bottomrule
    \end{tabular}
\end{table}

\subsection{Analysis}

To investigate how the model exploits acoustic and linguistic cues, we analyze encoder-level and temporal-level contributions on the Smart-Turn and in-house Japanese test sets, using a unified gradient-based attribution framework complemented by controlled encoder ablations.

\begin{figure}[t]
\centering
    \includegraphics[width=0.8\linewidth]{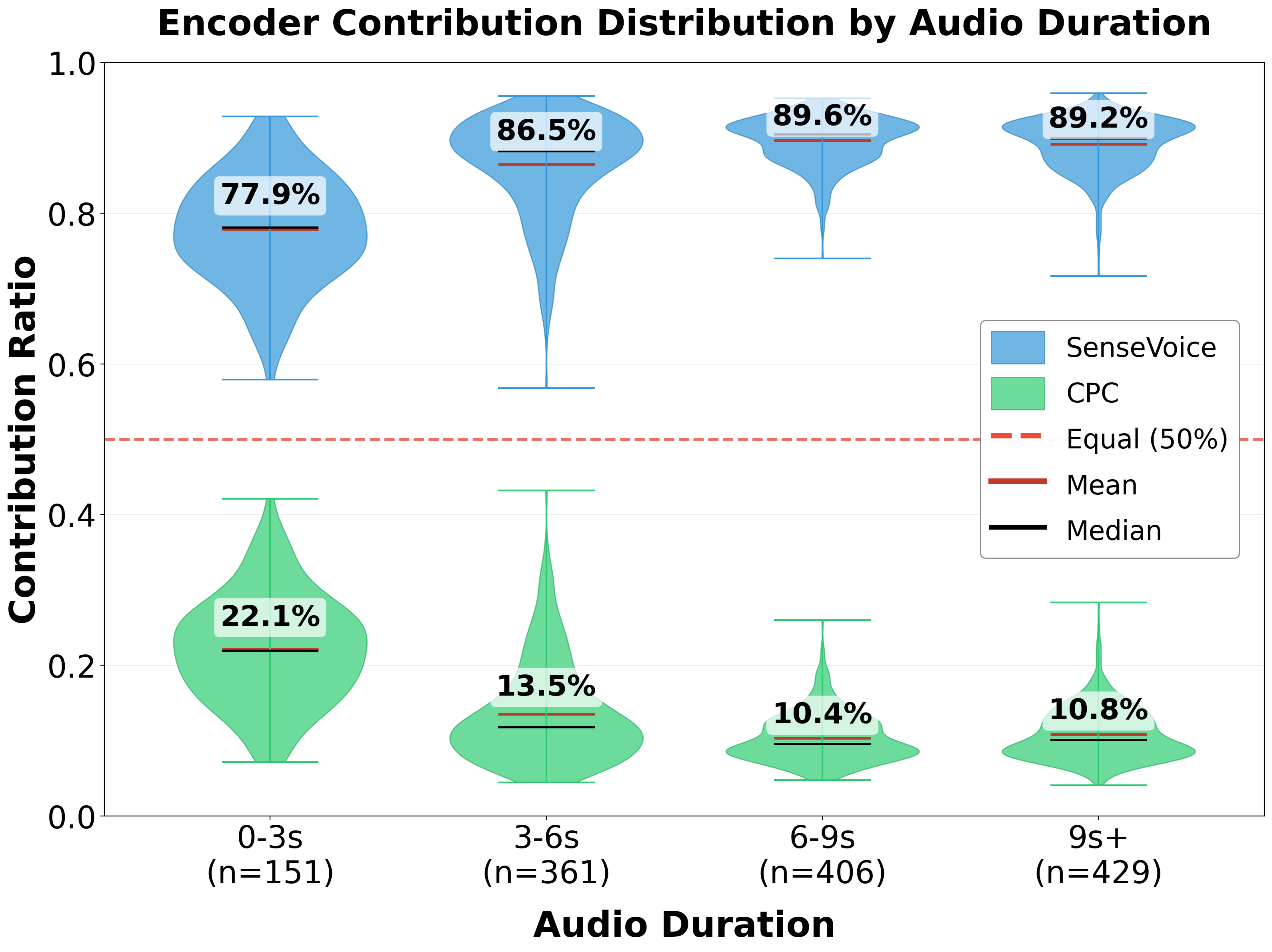}
    \caption{Encoder-wise contributions across utterance durations for SenseVoice and CPC.}
    \label{fig:ablation_study}
\end{figure}

\subsubsection{Encoder-Level Contribution.}

For each sample, we backpropagate the classification score to the encoder outputs and compute encoder-wise scores via the element-wise gradient–activation product. The contribution ratio for encoder $i$ is defined as
\[
\rho_i = \frac{\lVert \nabla_{x_i} \odot x_i \rVert_2}{\sum_j \lVert \nabla_{x_j} \odot x_j \rVert_2},
\]
where $x_i$ denotes the output features of encoder $i$. We group samples into four duration bins: 0–3 s, 3–6 s, 6–9 s, and $>$9 s.

As shown in Fig.~\ref{fig:ablation_study}, SenseVoice clearly dominates across all bins, with $\rho_{\text{Sense}}$ increasing from roughly 0.78 (0–3 s) to 0.90 (6–9 s), indicating that linguistically enriched representations are the primary driver of turn-taking decisions and become even more influential for longer contexts. CPC exhibits complementary but smaller contributions, with $\rho_{\text{CPC}}$ decreasing from about 0.22 (0–3 s) to 0.10 (6–9 s) as duration grows, suggesting that prosodic and low-level acoustic cues are most useful for short utterances when semantic context is limited. Overall, this reveals a length-dependent division of labor: SenseVoice provides the dominant semantic signal, while CPC supplies auxiliary acoustic evidence, particularly in short-context scenarios.

\begin{figure}[htbp]
\centering
    \includegraphics[width=\linewidth]{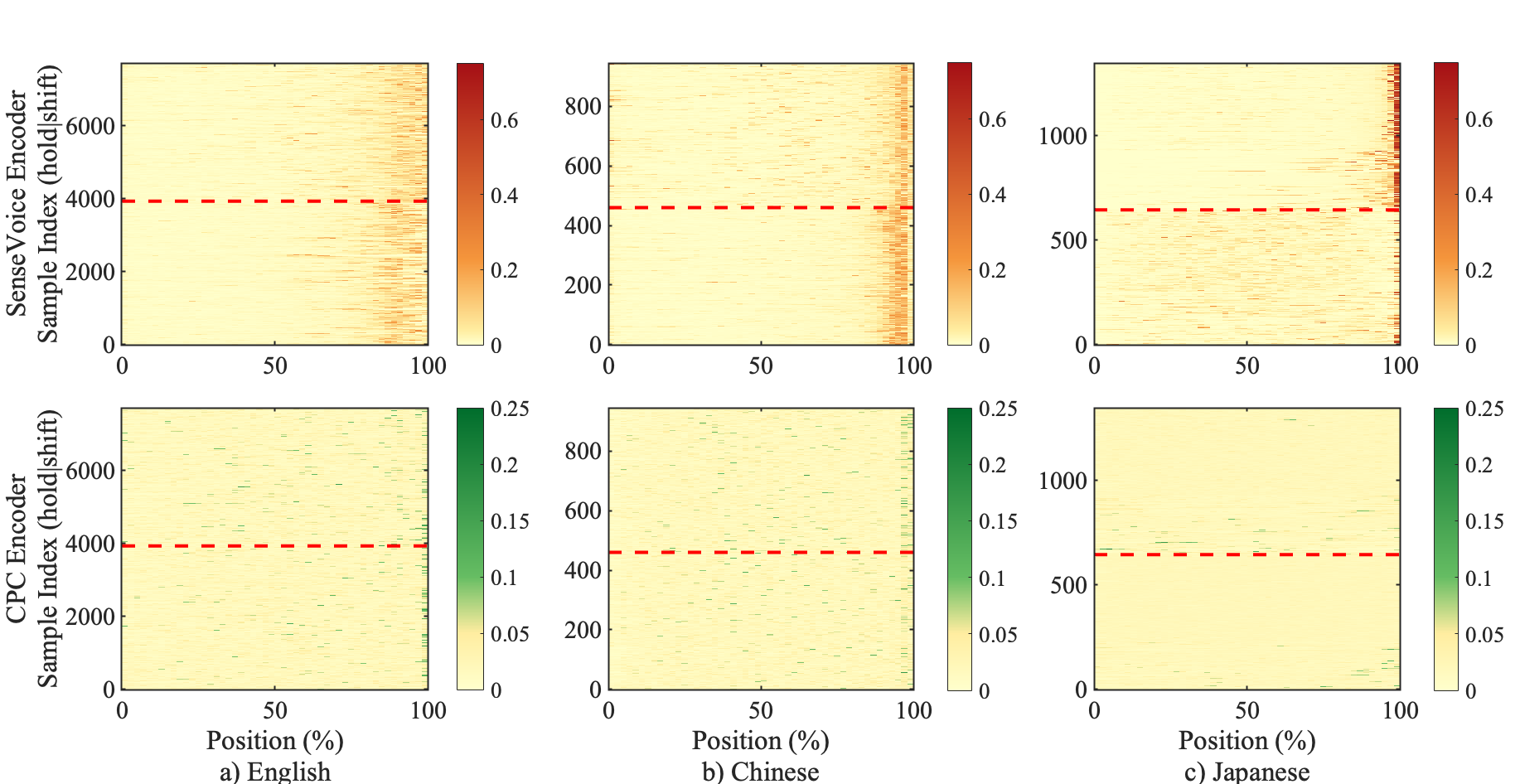}
    \caption{Temporal-level attribution heatmaps for the three languages (normalized to $0$--$100\%$). The upper and lower rows correspond to the SenseVoice and CPC encoders, respectively. Red dashed lines indicate the boundary between Hold and Shift samples.}
    \label{fig:temporal_heatmap}
\end{figure}

\subsubsection{Temporal-Level Contribution.}

To analyze how the model allocates attention over time, we perform \emph{position-regularized temporal attribution} via the gradient--input product, projecting temporal contribution scores onto a normalized $0$--$100\%$ axis to enable direct comparison across variable-length utterances.

Across all three languages, contributions from the SenseVoice encoder exhibit a consistent monotonic increase toward the end of the utterance, revealing a strong bias toward utterance-final regions for turn-taking prediction (Fig.~\ref{fig:temporal_heatmap}). However, the degree of temporal concentration varies substantially across languages. In Japanese, the majority of the SenseVoice contribution is concentrated within the final \textbf{5\%} of the audio context. In contrast, Chinese shows a broader concentration around the final \textbf{10\%}, while English displays a more gradual accumulation, with peak contributions occurring around the final \textbf{25\%} of the utterance.
This localization reflects language-specific turn-completion mechanisms. In Japanese, turn-taking cues are tightly tied to utterance-final phonetic structure and sentence-final morphemes such as polite-form endings (e.g., ``\textit{-masu}'', ``\textit{-desu}''), which are fully realized only in \textit{Shift} instances and thus yield highly concentrated end-of-utterance attention; English instead relies on anticipatory prosodic patterns unfolding over a longer span, with Chinese intermediate.
Notably, the CPC encoder shows no comparable temporal bias: its contribution stays roughly uniform across the utterance for all three languages, indicating that low-level acoustic representations capture global prosodic characteristics rather than temporally localized completion cues. Together, these findings suggest that SenseVoice, as an ASR encoder, models temporally localized, language-dependent turn signals through its rich phonetic, sub-lexical, and lexical information (e.g., sentence-final morphemes), whereas CPC provides complementary, globally distributed prosodic variation not fully captured by ASR-style encoders, enriching the representation along additional acoustic dimensions.

\section{Conclusion}

We presented JAL-Turn, a lightweight and robust turn-taking framework for full-duplex spoken dialogue systems. By jointly modeling acoustic and linguistic cues via cross-attention fusion, a lightweight Transformer, and temporal attention pooling—with a frozen ASR encoder shared for parallel inference—it delivers low-latency, accurate turn-taking prediction, while a scalable pipeline automatically derives reliable labels from large-scale real-world corpora. Across multilingual public benchmarks and an in-house Japanese dataset, JAL-Turn matches or surpasses strong audio-only and LLM-based baselines in detection quality, and remains competitive with substantially heavier SLM-based systems, while achieving lower latency suitable for real-time full-duplex deployment.

\section{Limitations}
Our work has several limitations. First, the data pipeline derives supervision from stereo VAD with a future-window rule (\(\sim\)85\% estimated labeling accuracy), so it is noisy and only indirectly reflects pragmatic intent---ambiguous overlaps, long hesitations, or VAD errors may mislabel hold/shift---and its explicit suppression of backchannels partly explains the backchannel gap on Easy-Turn. Second, JAL-Turn targets binary hold/shift prediction over a fixed 10\,s context; while effective for low-latency deployment, this does not capture richer turn-management behaviors (e.g., multi-state dialog acts or explicit backchannel prediction), which we leave to future work. Third, part of our evaluation uses an in-house Japanese corpus with a manually labeled test subset, which may limit full reproducibility and independent verification.

%
%
%
\begin{credits}
\subsubsection{\ackname}
This work was supported by Recho, Inc.

\subsubsection{\discintname}
All authors are employees of Recho, Inc. and have no other competing interests to declare.
\end{credits}

\section*{Appendix}

\noindent\textbf{Prompt for LLM-based turn-taking detection.}
\begin{quote}\small\ttfamily

You are analyzing a conversation turn. Your task is to determine whether the
current speaker should continue speaking or should stop and let the other
speaker talk.\\[2pt]
Rules: If the current input text indicates that the speaker has NOT finished
their thought or sentence (the speech is incomplete or continuing), output:
hold. If the current input text indicates that the speaker HAS finished their
thought or sentence (the speech is complete and ready for the other speaker to
respond), output: shift.\\[2pt]
Analyze the following text and output your decision.
\end{quote}

\bibliographystyle{splncs04}
\bibliography{custom}

\end{document}